\documentclass[runningheads]{llncs}
\pdfoutput=1

\usepackage{graphicx}
\usepackage{amsmath,amssymb} % define this before the line numbering.
\usepackage{color}

\usepackage{stmaryrd}

\usepackage{algorithm}
\usepackage{booktabs}
\usepackage[colorlinks=true,urlcolor=blue,citecolor=blue,linkcolor=blue,bookmarks=true]{hyperref}
\usepackage{siunitx}
\usepackage{hyperref}

\usepackage{multirow}
\usepackage{cite}

\usepackage{tikz}

\usepackage{subcaption}

\usetikzlibrary{positioning}

\begin{document}
\pagestyle{headings}
%\mainmatter
\def\ACCV22SubNumber{268}  % Insert your submission number here

\title{Enhancing Fairness of Visual Attribute Predictors} % Replace with your title

%\titlerunning{ACCV-22 submission ID \ACCV22SubNumber}

%\authorrunning{ACCV-22 submission ID \ACCV22SubNumber}

\author{Tobias Hänel\inst{1} \and
Nishant Kumar\inst{1} \and Dmitrij Schlesinger\inst{1} \and Mengze Li\inst{2} \and Erdem Ünal\inst{1} \and Abouzar Eslami\inst{2}
\and Stefan Gumhold\inst{1}}

\institute{$^{1}$Chair for Computer Graphics and Visualization, TU Dresden, Germany \\ 
$^{2}$Carl Zeiss Meditec AG, Munich, Germany \\ 
Corresponding author: \email{tobias.haenel@tu-dresden.de}}

\maketitle
%===========================================================
\begin{abstract}
The performance of deep neural networks for image recognition tasks such as predicting a smiling face is known to degrade with under-represented classes of sensitive attributes. We address this problem by introducing fairness-aware regularization losses based on batch estimates of Demographic Parity, Equalized Odds, and a novel Intersection-over-Union measure. The experiments performed on facial and medical images from CelebA, UTKFace, and the SIIM-ISIC melanoma classification challenge show the effectiveness of our proposed fairness losses for bias mitigation as they improve model fairness while maintaining high classification performance. To the best of our knowledge, our work is the first attempt to incorporate these types of losses in an end-to-end training scheme for mitigating biases of visual attribute predictors. Our code is available at \href{https://github.com/nish03/FVAP}{this https URL}.
\end{abstract}

\section{Introduction}
\label{sec:intro}

The manifestation of bias is evident in every aspect of our society, from educational institutions \cite{ref44}, to bank credit limits for women \cite{ref45}, to criminal justice for people of color \cite{ref46}. The core problem is the inability of an individual to make ethically correct objective decisions without being affected by personal opinions. With the advent of recent machine learning (ML) algorithms trained on big data, there is a dramatic shift towards using such algorithms to provide greater discipline to impartial decision-making. However, ML-based algorithms are also prone to making biased decisions \cite{ref22,ref47}, as the reliability of data-based decision-making is heavily dependent on the data itself. For instance, such models are unfair when the training data is heavily imbalanced towards a particular class of a sensitive attribute such as race \cite{ref34}. A notable example by \cite{ref22} shows that by assuming the ML model’s target attribute as gender and the sensitive attribute as skin color, the classification error rate is much higher for darker females than for lighter females. A similar concern exists in the medical fraternity, where a recent work \cite{ref78} studied the correlation between the under-representation of darker skin images with the classification error in predicting dermatological disease. Another work \cite{ref79} showed a sharp decrease in classification performance while diagnosing several types of thoracic diseases for an under-represented gender in the X-ray image training data set. Therefore, it is critical to mitigate these biases in ML-based models for visual recognition tasks, to alleviate ethical concerns while deploying such models in real-world applications.  
   
\begin{figure}[!t]
\centering
\includegraphics[width=120mm]{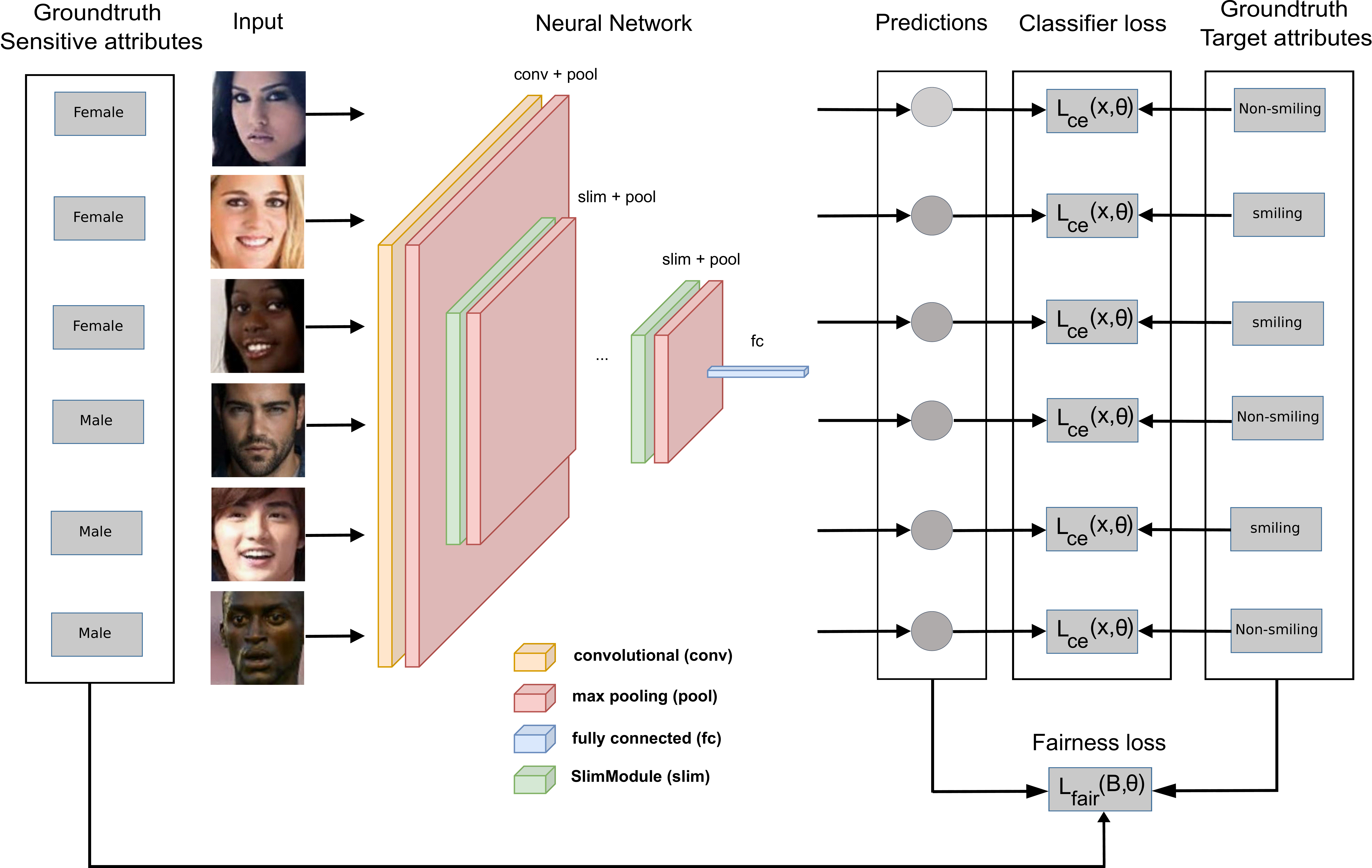} 
\caption{The figure shows our new training procedure that improves the fairness of image classification models w.r.t.~to sensitive attributes such as gender, age, and ethnicity. We add a weighted fairness loss to the standard cross-entropy loss during mini-batch gradient descent. It estimates the fairness of the model's predictions based on the sensitive attributes and the predicted and ground-truth target attributes from the samples within each batch.}
\label{fig:ACCV22}
\end{figure}

Recent studies focus on creating balanced data sets \cite{ref10}, or perform data augmentation \cite{rotemberg_patient-centric_2020} to remove imbalance with respect to the sensitive attributes. We argue that to make an ML model useful in real-life scenarios, it should achieve algorithmic fairness while still being trained on data sets that consist of real-world biases. In terms of algorithmic fairness, works such as \cite{ref5,ref6,ref27} aim to learn the features in the data that are statistically independent of the sensitive attributes, while \cite{ref9,ref48} focus on de-biasing the latent space of a generative model to achieve a fair outcome. We propose that to reduce bias w.r.t.~the sensitive attributes, a model must satisfy the fairness notations by learning them during training. Additionally, none of the previous approaches attempted to utilize an inherent IoU-based fairness measure to train an ML model and achieve algorithmic fairness without loss in classification accuracy. Our overall learning scheme is presented in Fig.~\ref{fig:ACCV22}. The contributions of our work are as follows:

\begin{itemize}
\item We use classical fairness notations such as Demographic Parity (DP) and Equalized Odds (EO) to define the corresponding fairness loss terms and measure the deviations from the assumptions of the probabilistic independence w.r.t.~sensitive attributes. We quantify these deviations by using mean squared error as well as the Kullback-Leibler divergence (KLD) between the learned probability distribution and the best-factorized distribution, leading to the mutual information (MI) between the variables in the learned model.

\item We generalize the fairness notations such as DP and EO for categorical variables since the task is usually a multi-class problem in image classification. In the past, such notations were defined for binary variables only. 

\item We introduce a novel fairness loss based on the Intersection-over-Union(IoU) measure and study its relevancy for achieving fair classification results empirically. Our experiments show that it can simultaneously improve the model fairness and the baseline classification performance when the model is evaluated with fairness metrics. 
    
\item We exhaustively evaluate all introduced losses with facial attribute prediction on CelebA \cite{ref35}, age group estimation on UTKFace \cite{ref41}, and disease classification on the SIIM-ISIC Melanoma data set \cite{rotemberg_patient-centric_2020}. It was possible for all of these data sets to improve the model fairness with our method.

\end{itemize}

\section{Related Work}
We discuss current methods that deal with bias mitigation in the data domain and provide an overview of works conducted to achieve fairness-aware facial and medical image recognition systems. 

\paragraph{\normalfont\textbf{Mitigating bias in the data domain:}} The work  \cite{ref23} developed an audit process to highlight the fairness-driven concerns in facial recognition while \cite{ref27} provided a benchmark for bias removal and highlighted key ethical questions for mitigating bias in vision data sets. Another study \cite{ref78} estimated skin tones for images with skin disease and showed that the darker skin population is under-represented in the data set. In \cite{ref30}, the authors performed data augmentation to mitigate bias in facial image data sets. Multiple authors \cite{ref10,ref67} presented new facial data sets with a balanced number of samples from different classes of sensitive attributes. The works such as \cite{ref65,ref85,ref15} used image generation models to synthesize new facial images and used these images with standard pre-trained classifiers to investigate algorithmic performance bias based on sensitive attributes. Some authors \cite{ref4,ref48} used the latent space of the Generative Adversarial Networks (GAN) to de-correlate the target and sensitive attributes, while \cite{ref26} developed a tool that uses an image data set and its annotations as input and suggests actions to mitigate potential biases. The work \cite{ref6} learned a GAN-based mapping of the source image to a target fair image representation by removing the semantics of target features such as eyes and lips from the sensitive attribute gender.
%%%%%%%%%%%%% Add Rebuttal 3%%%%%%%%%%%%%%%%
The issue of unfair predictions due to biased data also exists in few-shot \cite{few-shot} and zero-shot \cite{zero-shot} learning where \cite{few-shot} includes a constraint called  decision boundary covariance, enabling independence of target and sensitive attributes while \cite{zero-shot} maps the unlabelled target classes to a point in semantic space exterior to the point cluster of labeled source classes. 

\paragraph{\normalfont \textbf{Fairness in facial recognition:}}
A study \cite{ref97} proposed a classifier that predicts target facial attributes using pre-learned sensitive attributes. Another work \cite{ref110} calibrated the predicted target class labels to reduce the performance bias while \cite{ref96} proposed separate fair classifiers for each target class. In \cite{ref98}, they used the cross-entropy loss between the predicted sensitive labels and the uniform distribution, causing the model to be confused and invariant to the sensitive attributes. Many authors \cite{ref70,ref111,ref9,ref99,ref109,ref69} proposed adversarial learning to obtain a fair ML model. A study \cite{ref106} presented channel-wise attention maps that adapt to diversity in facial features in every demographic group to obtain a fair classifier. A publication \cite{ref101} used Q-learning to train an agent by rewarding less skew of the inter-class distance between the sensitive attribute such as race. In \cite{ref104}, they used a simple encoder-decoder network, \cite{ref5,ref6} used adversarial learning of GAN, and \cite{ref63} used a style transfer GAN to synthesize facial images independent of sensitive attributes such as gender and race. A work \cite{ref105} ensured that the generated images from an image generation model have the same error rates for each class of the sensitive attributes. The authors of \cite{ref107} introduced the False Positive Rate (FPR) as the penalty loss to mitigate bias. The authors in \cite{ref29} showed that adversarial learning might worsen classification accuracy and fairness performance. They suggested a constraint that standard classification accuracy and fairness measures should be limited to the average of both metrics.

\paragraph{\normalfont \textbf{Fairness in disease diagnosis:}}
 In \cite{ref79}, they showed the problem of gender imbalance in medical imaging data sets, where the classification performance is less for disease diagnosis of the under-represented gender. A study \cite{ref116} found that True Positive Rate (TPR) disparity for sensitive attributes such as gender, age, and race exists for all classifiers trained on different data sets of chest X-ray images. Another publication \cite{ref117} highlighted some recent works to mitigate bias via federated learning (FL) \cite{ref118,ref119,ref120}. FL enables multiple clients stationed at different geographical locations to train a single ML model with diversified data sets collaboratively. This method should help overcome the under-representation of individual classes of sensitive attributes in the data, resulting in unbiased models. However, data heterogeneity among the distributed clients remains a challenge for such FL-based models, which might degrade the performance. 

In summary, past works did not thoroughly explore the theoretical insights into fairness-based regularization measures. They also did not apply them in experiments on visual data to obtain a model that has high classification performance on target attributes and is unbiased w.r.t.~sensitive attributes.

\section{Method}
Let $x\in\cal X$ denote an image and $T=(x_1, x_2, \ldots, x_{|T|})$ be the training data consisting of $|T|$ images. Let $y_s\in\{1\ldots K_s\}$ and $y_t\in\{1\ldots K_t\}$ be {\em sensitive} and {\em target} attributes respectively. The latter is our classification target, whereas the former is the one, classification should not depend on. The asterisk will be used to denote the attribute ground truth values, e.g., ~$y_s^*(x)$ means the ground-truth value of the sensitive attribute for an $x$ from $T$. We treat a trainable {\em classifier} as a conditional probability distribution $p_\theta(y_t|x)$ parameterized by an unknown parameter $\theta$ to be learned. In our work, we use Feed Forward networks (FFNs) as classifiers, which means $\theta$ summarizes network weights.

Next, we consider the joint probability distribution $p(x,y_t)=p(x)\cdot p_\theta(y_t|x)$, where $p(x)$ is some probability distribution, from which training data $T$ is drawn. Noteworthy, the ground truths can also be considered as random variables, since they are deterministic mappings from $\cal X$, i.e., there exists a unique value $y_t^*(x)$ as well as a unique value $y_s^*(x)$ for each $x\in\cal X$. Hence, for the ground truths, we can also consider their statistical properties like joint probability distribution $p(y^*_s, y_t)$, independence $y_t \perp y_s^*$, or similar.

\subsubsection{Demographic Parity.}
The notation reads $y_t \perp y_s^*$, i.e.~the prediction should not depend on the sensitive attribute. Traditionally (for binary variables) a classifier is said to satisfy demographic parity if
\begin{equation}\label{eq:dp_bin}
    p(y_t=1|y^*_s=1)=p(y_t=1)
\end{equation}
holds. A straightforward generalization to the case of categorical variables is to require the same for all possible values, i.e.
\begin{equation}\label{eq:dp}
    p(y_t=a|y^*_s=b)=p(y_t=a) ,
\end{equation}
where $a\in\{1\ldots K_t\}$ and $b\in\{1\ldots K_s\}$.
The above notation can be used to define a {\em loss} function, i.e.~measure that penalizes the deviation of a given probability distribution $p_\theta$ from satisfying \eqref{eq:dp}. One possible way is to penalize the sum of squared differences
\begin{equation}\label{eq:dp_losssqrt}
L^{l_2}_{dp}(\theta) = \sum_{ab} \bigl[p_\theta(y_t=a|y^*_s=b)-p_\theta(y_t=a)\bigr]^2 .
\end{equation}

In fact, during the transition from \eqref{eq:dp} to \eqref{eq:dp_losssqrt} we compare the actual joint probability distribution $p_\theta(y_t,y^*_s)$ to the corresponding factorized (i.e.~independent) probability distribution $p_\theta(y_t)\cdot p_\theta(y^*_s)$ using squared difference, i.e., interpreting probability distributions as vectors to some extent\footnote{Strictly speaking, it directly holds only if $p_\theta(y^*_s)$ is uniform, otherwise \eqref{eq:dp_losssqrt} corresponds to a squared difference between $p_\theta(y_t,y^*_s)$ and $p_\theta(y_t)\cdot p_\theta(y^*_s)$, where addends are additionally weighted by $1/p_\theta(y^*_s)^2$.}. The squared difference is however not the only way to compare probability distributions. Another option would be e.g., Kullback-Leibler divergence $D_{KL}(p_\theta(y_t,y^*_s)||p_\theta(y_t)\cdot p_\theta(y^*_s))$ which leads to the mutual information loss
\begin{eqnarray}\label{eq:dp_lossmi}
    L^{mi}_{dp}(\theta) & = & \sum_{ab}p_\theta(y_t=a, y^*_s=b)\cdot
    \log\frac{p_\theta(y_t=a, y^*_s=b)}{p_\theta(y_t=a)\cdot p_\theta(y^*_s=b)} = \nonumber \\
    & = & H(y_t) + H(y^*_s) - H(y_t, y^*_s) ,
\end{eqnarray}
where $H(\cdot)$ denotes the entropy. Note that we derived different losses \eqref{eq:dp_losssqrt} and \eqref{eq:dp_lossmi} from the same independence requirements \eqref{eq:dp} using different distance measures for probability distributions.

\subsubsection{Equalized Odds.}
It is assumed that the predicted target attribute and the ground truth sensitive attribute are {\em conditionally} independent given a fixed value of the ground truth target, i.e.~$(y_t \perp y_s^*)|y_t^*$. Hence
\begin{equation}\label{eq:eo}
    p(y_t=a|y^*_t=b,y^*_s=c)=p(y_t=a|y^*_t=b)
\end{equation}
should hold for all triples $a, b\in\{1\ldots K_t\}$ and $c\in\{1\ldots K_s\}$. Again, similarly to the previous case, we consider first the simple quadratic loss
\begin{equation}\label{eq:eo_losssqrt}
L^{l_2}_{eo}(\theta) = \sum_{abc} \bigl[p(y_t=a|y^*_t=b,y^*_s=c) - p(y_t=a|y^*_t=b) \bigr]^2 .
\end{equation}
It is also possible to measure the deviations of the current model $p_\theta(y_t,y^*_t,y^*_s)$ from the requirements \eqref{eq:eo} utilizing the corresponding mutual information as 
\begin{equation}\label{eq:eo_lossmi}
     L^{mi}_{eo}(\theta) = \sum_a\Bigl[ H(y_t|y^*_t=a) + H(y^*_s|y^*_t=a) - H(y_t, y^*_s|y^*_t=a) \Bigr] .
\end{equation}

\subsubsection{Intersection-Over-Union.}
The above losses have a distinct statistical background since they rely on specific independence assumptions. In practice, however, we are often not interested in making some variables independent. Instead, the general goal could be phrased as ``the classification performance should be similar for different values of the sensitive attribute''. Hence, the core question is how to measure classifier performance adequately. We argue for the IoU measure because it can appropriately rate performance, especially for unbalanced data.
For a target value $a$, the corresponding IoU is traditionally defined as
\begin{equation}\label{eq:iou1}
    \text{IoU}_\theta(a) = \frac{p_\theta(y_t=a \wedge y^*_t=a)}{p_\theta(y_t=a \vee y^*_t=a)} ,
\end{equation}
where $\wedge$ and $\vee$ denote logical ``and'' and ``or'' respectively. The overall IoU is usually defined by averaging \eqref{eq:iou1} over $a$, i.e.
\begin{equation}\label{eq:iou1_av}
    \text{IoU}_\theta = \frac{1}{K_t}\sum_a \text{IoU}_\theta(a) .
\end{equation}
 For a model $p_\theta(y_t,y^*_t,y^*_s)$ with target value $a$ and  sensitive value $b$, we extend \eqref{eq:iou1} and define $\text{IoU}_\theta(a,b)$ as
\begin{equation}\label{eq:iou2}
    \text{IoU}_\theta(a,b) = \frac{p_\theta((y_t=a \wedge y^*_t=a) \wedge y^*_s=b)}
    {p_\theta((y_t=a \vee y^*_t=a) \wedge y^*_s=b )} .
\end{equation}
Now given \eqref{eq:iou1_av}, $\text{IoU}_\theta(b)$ for a particular value $b$ of the sensitive attribute is
\begin{equation}\label{eq:iou2_av}
    \text{IoU}_\theta(b) = \frac{1}{K_t}\sum_a \text{IoU}_\theta(a,b) 
\end{equation}
and the loss penalizes the deviations of these IoU-s from the overall IoU \eqref{eq:iou1_av} as:
\begin{equation}\label{eq:iou_losssqrt}
    L_{iou}(\theta) = \sum_b \Bigl[\text{IoU}_\theta(b)-\text{IoU}_\theta\Bigr]^2 .
\end{equation}
Note that to define \eqref{eq:iou1} to \eqref{eq:iou_losssqrt}, we again used statistical interpretation of all involved variables, i.e., the joint probability distribution $p_\theta(y_t,y^*_t,y^*_s)$. This time however we do not explicitly enforce any independence in contrast to the demographic parity or equalized odds.

\subsubsection{Optimization.}
All introduced losses are differentiable w.r.t.~unknown parameter $\theta$\footnote{We also assume that probabilities are differentiable w.r.t.~parameters.} because we use probability values for their computation. Hence, we can mix them with other losses simply and conveniently. In particular, if the classifier is e.g., an FFN, we can optimize them using error back-propagation. Second, it should be noted that all losses rely on low-order statistics, i.e., it is only necessary to estimate current $p_\theta(y_t,y^*_t,y^*_s)$ to compute them. For example, if the involved variables are binary, we only need to estimate  $8$ values. We assume that they can be reliably estimated from a data mini-batch of a reasonable size instead of computing them over the whole training data. This makes optimizing the introduced losses within commonly used stochastic optimization frameworks possible. To be more specific, the overall loss can be written as
\begin{eqnarray}\label{eq:loss_total}
    L(\theta) & = & \sum_{x\in T}L_{ce}(x,\theta) \ + \ \lambda\cdot L_{fair}(T,\theta) =  \nonumber \\
    & = & \mathbb E_{B\subset T} \sum_{x\in B}L_{ce}(x,\theta) \ + \ \lambda\cdot L_{fair}(T,\theta) ,
\end{eqnarray}
where the expectation $\mathbb E$ is over all mini-batches $B\subset T$ randomly sampled from the training data, $L_{ce}$ is a ``usual'' classification loss, e.g., Cross-Entropy, $\lambda$ is a weighting coefficient, and $L_{fair}$ is one of the fairness losses introduced above. For the sake of technical convenience, we approximate \eqref{eq:loss_total} by
\begin{equation}\label{eq:loss_final}
    L(\theta) = \mathbb E_{B\subset T} \left[\sum_{x\in B}L_{ce}(x,\theta) \ + \ \lambda\cdot L_{fair}(B,\theta)\right] .
\end{equation}

\subsubsection{Impact on the Performance.}
We want to show a crucial difference between demographic parity and equalized odds. Imagine a hypothetical ``perfect classifier'' that always assigns probability $1$ to the ground truth label. Hence, the requirements of demographic parity $y_t \perp y_s^*$ turn into $y_t^* \perp y_s^*$. It means that the perfect classifier can satisfy demographic parity only if the ground truth target and the ground truth sensitive attributes are completely uncorrelated, which is hard to expect in practice. It follows from the practical perspective that the classifier performance should decrease when we try to make the classifier fair in the sense of demographic parity.

On the other side, in our notations, the case of a perfect classifier can be written as
\begin{equation}
p_\theta(y_t,y^*_t,y^*_s)=p_\theta(y_t|y^*_t,y^*_s) \cdot p(y^*_t,y^*_s) = \llbracket y_t=y^*_t\rrbracket\cdot p(y^*_t,y^*_s) ,
\end{equation}
where $\llbracket\cdot\rrbracket$ is $1$ if its argument is true. Hence, without loss of generality
\begin{equation}
p_\theta(y_t,y^*_t,y^*_s)=\llbracket y_t=y^*_t\rrbracket\cdot p(y^*_t,y^*_s) = p_\theta(y_t|y^*_t) \cdot p(y^*_s|y^*_t)\cdot p(y^*_t) .
\end{equation}
It means that for a perfect classifier, the predicted target attribute and ground-truth sensitive attribute are conditionally independent, i.e., a perfect classifier automatically satisfies the requirements of equalized odds $(y_t \perp y_s^*)|y_t^*$. In practice, if the baseline classifier is already good enough, its performance should not worsen when we try to make the classifier fair w.r.t.~equalized odds. 

Considering the IoU-loss \eqref{eq:iou_losssqrt}, it is easy to see that it is zero for a perfect classifier, just because all IoU values are equal to one in this case. Hence, as in the case of equalized odds, we do not expect a drop in classifier performance when we try to make it fair w.r.t.~the IoU-loss.

\subsubsection{Fairness and Calibration Properties.}
We consider a {\em linear squeezing} operation as follows. Let $y\in\{0,1\}$ be a binary variable\footnote{We discuss in detail only the case of binary variables and the IoU-loss for simplicity. The situation is similar for other cases.} and $p(y|x)$ a conditional probability distribution for an input $x$. The linear squeezing puts all probability values into the range $[0.5-\alpha/2,0.5+\alpha/2] $ with $0 < \alpha < 1$ by applying 
\begin{equation}\label{eq:squeezing}
    p'(y|x) = \bigl[p(y|x) - 0.5\bigr]\cdot\alpha + 0.5 .
\end{equation}
Firstly, this operation does not change the {\em decision} about $y$ for a given $x$. The decision is made by thresholding $p(y|x)$ at the $0.5$ level, which does not change after applying the linear squeezing. Secondly, it makes the classifier ``less confident'' about its decision because the output probabilities lie in a narrower range.
At the same time, linear squeezing can be understood as mixing the original $p(y|x)$ and uniform distribution, since \eqref{eq:squeezing} can be rewritten as
\begin{equation}\label{eq:mixing}
    p'(y|x) = p(y|x)\cdot\alpha + 0.5\cdot (1-\alpha).
\end{equation}
Consider now the confusion matrix, i.e.~$p(y_t, y_t^*)$, obtained by averaging over the training data, and the corresponding IoU-value \eqref{eq:iou1_av} (for now we do not consider the sensitive attribute). Let us assume evenly distributed ground truth labels for simplicity. So applying \eqref{eq:mixing} to the output probabilities gives
\begin{equation}\label{eq:mixing1}
    p'(y_t, y_t^*) = p(y_t, y_t^*)\cdot\alpha + 0.25\cdot (1-\alpha).
\end{equation}
For $\alpha$ close to $1$, the IoU obtained from $p'(y_t, y_t^*)$ (i.e., squeezed by \eqref{eq:mixing1}) will be close to the original IoU (i.e.~obtained from the original non-squeezed $p(y_t, y_t^*)$) for which we assume a rather high value since the classifier is essentially better than random chance. For $\alpha$ close to zero, the modified IoU converges to $1/3$. Hence, IoU differences (i.e.~addends in \eqref{eq:iou_losssqrt}) vanish. To conclude, we can make the IoU-loss \eqref{eq:iou_losssqrt} alone arbitrarily small just by applying linear squeezing without changing the decision rule.

Note that the squeezing operation with a small $\alpha$ makes the primary loss, i.e., the cross-entropy, essentially worse since the log-likelihoods of the ground truth labels get smaller. In fact, the fairest classifier is a random choice decision, i.e., which does not depend on the input. It is fair and under-confident but poor in terms of the primary goal, i.e.~classification accuracy, and w.r.t.~the primary cross-entropy loss.
To conclude, adding the IoU-loss to the primary objective (see \eqref{eq:loss_final}) pushes the solution towards being less confident. It may be a desired or an undesired behavior depending on whether the baseline classifier is already well-calibrated or not. For over-confident baseline classifiers, employing the IoU-loss should improve calibration properties. The calibration properties may get worse for already well-calibrated or under-confident baseline classifiers.

\section{Experiments}
We validate our contributions on three data sets. The first study of interest in \ref{sec:celeba} concerns the CelebFaces attributes (CelebA) data set \cite{ref35} which contains more than 200K images of celebrities and manually annotated facial attributes. Secondly in \ref{sec:utkface}, we investigate the UTKFace facial image data set \cite{ref41} which contains over 20K images from a broad range of age groups. In addition to the facial images, we perform experiments with a data set from the SIIM-ISIC Melanoma classification challenge \cite{rotemberg_patient-centric_2020} in \ref{sec:mela} that contains 33k+ skin lesion images. We focus on achieving a balanced target attribute prediction performance that does not depend on the sensitive attribute. We split each data set into a train, validation, and test partition to verify the results of our method. First, we train a baseline model (details in the supplementary material). To improve its fairness, we continue the optimization process by extending the cross-entropy loss with one of the weighted fair losses (see \ref{eq:loss_final}) and perform experiments with two different strategies for selecting $\lambda$ in \ref{sec:lambda}. 

%\paragraph{\normalfont\textbf{CelebA}}
\subsection{CelebA}
\label{sec:celeba}
For experiments with the CelebA data set, we use SlimCNN \cite{sharma_slim-cnn_2020}, a memory-efficient convolutional neural network, to predict whether a depicted person is smiling or not. To evaluate how our method influences the fairness of this prediction task, we select the binary variables $Male$ and $Young$ (representing gender and age) as sensitive attributes $y_s$. We use the original train, validation, and test partitions in all experiments with this data set.

\begin{table*}[!t]
	\centering
	\caption{Results of the experiments with manually selected weighing coefficients $\lambda$ for CelebA facial attribute prediction. The task is to predict the binary target attribute $y_t = Smiling$. The experiments \#1 to \#6 use the sensitive attribute $y_s = Male$ for the evaluation of the fairness loss terms, while \#7 to \#12 use the sensitive attribute $y_s = Young$.  The values in bold are the best results of each evaluation metric.}
	\vspace{5pt}
	\begin{tabular}{|*{8}{c|}}
		\hline
		\scriptsize\# & Loss           & $Acc$   &   $L_{iou}$ \eqref{eq:iou_losssqrt}  & $L^{l_2}_{eo}$ \eqref{eq:eo_losssqrt} & $L^{mi}_{eo}$ \eqref{eq:eo_lossmi} & $L^{l_2}_{dp}$ \eqref{eq:dp_losssqrt} & $L^{mi}_{dp}$ \eqref{eq:dp_lossmi}  \\ \hline
		\scriptsize1  & $L_{ce}$       & 0.902   & \num{8.73E-04} & \num{4.89E-03} & \num{5.12E-03} & \num{1.77E-02} & \num{8.46E-03} \\
		\scriptsize2  & $L_{iou}$      & $\mathbf{0.903}$ & \num{7.32E-05} & \num{8.59E-04} & \num{4.26E-04} & \num{2.51E-03} & \num{1.20E-03} \\
		\scriptsize3  & $L^{l_2}_{eo}$ & 0.902   & $\mathbf{1.35 \times 10^{-5}}$ & $\mathbf{1.78 \times 10^{-4}}$ & $\mathbf{7.71 \times 10^{-5}}$ & $\mathbf{1.36 \times 10^{-4}}$ & $\mathbf{6.45 \times 10^{-5}}$ \\
		\scriptsize4  & $L^{mi}_{eo}$  & 0.899   & \num{2.37E-05} & \num{2.24E-04} & \num{1.03E-04} & \num{8.40E-04} & \num{4.00E-04} \\
		\scriptsize5  & $L^{l_2}_{dp}$ & 0.899   & \num{4.28E-04} & \num{3.75E-03} & \num{1.87E-03} & \num{1.57E-04} & \num{7.43E-05} \\
		\scriptsize6  & $L^{mi}_{dp}$  & 0.901   & \num{5.28E-04} & \num{7.73E-03} & \num{3.96E-03} & \num{7.15E-04} & \num{3.40E-04} \\ \hline 
		\scriptsize7  & $L_{ce}$       & 0.901   & \num{1.34E-03} & \num{8.28E-03} & \num{4.93E-03} & \num{1.16E-02} & \num{3.48E-03} \\
		\scriptsize8  & $L_{iou}$      & 0.901   & \num{4.15E-05} & \num{9.96E-04} & \num{3.06E-04} & \num{1.31E-03} & \num{3.96E-04} \\
		\scriptsize9  & $L^{l_2}_{eo}$ & $\mathbf{0.902}$   & $\mathbf{1.01 \times 10^{-5}}$ & $\mathbf{5.48 \times 10^{-5}}$ & $\mathbf{1.65 \times 10^{-5}}$ & $\mathbf{6.81 \times 10^{-5}}$ & $\mathbf{2.08 \times 10^{-5}}$ \\
		\scriptsize10 & $L^{mi}_{eo}$  & 0.901   & \num{1.64E-05} & \num{2.57E-04} & \num{7.40E-05} & \num{4.24E-04} & \num{1.29E-04} \\
		\scriptsize11 & $L^{l_2}_{dp}$ & 0.901   & \num{8.34E-04} & \num{6.93E-03} & \num{2.18E-03} & \num{1.58E-04} & \num{4.78E-05} \\
		\scriptsize12 & $L^{mi}_{dp}$  & 0.901   & \num{5.63E-04} & \num{4.26E-03} & \num{1.33E-03} & \num{1.45E-04} & \num{4.44E-05} \\ \hline
	\end{tabular}

	\label{tab:celeba}
\end{table*}

The results for the experiments with the best $\lambda$ values are shown in Table \ref{tab:celeba} (details in the supplementary material). Each row shows the results from an experiment with a particular training loss. The columns list the corresponding prediction accuracy ($Acc$) and all fairness metrics on the validation partition. The model fairness improved for all experiments with this data set according to almost all proposed losses. The application of the fairness losses $L_{iou}$ and the $L^{l_2}_{eo}$ did not lead to a reduction in the prediction accuracy. The $L^{l_2}_{eo}$ loss yielded the best fairness improvements according to all proposed fairness losses, while the $L_{iou}$ loss could even improve the classification performance. However, the model training with other losses slightly decreased the classification accuracy. Furthermore, the $L^{mi}_{eo}$ loss could improve model fairness w.r.t~ to all evaluated metrics, while the DP-based $L^{l_2}_{dp}$ and $L^{mi}_{dp}$ losses could only improve their respective fairness losses. Next, applying any fairness loss did not deteriorate the model performance with the sensitive attribute $y_s = Young$. 

%\paragraph{\normalfont\textbf{UTKFace}}
\subsection{UTKFace}
\label{sec:utkface}
 The images in UTKFace have annotations of a binary gender variable (Female, Male), a multi-class categorical ethnicity variable (White, Black, Asian, Indian, and Others), and an integer age variable (0-116 Years). Commonly this data set is used to perform age regression. We derive a categorical age group variable (under 31 Years, between 31-60 Years, over 60 Years) from the original ages as our predicted target attribute. $y_s = Ethnicity$ and $y_s = Gender$ represent the sensitive attributes in the experiments. We quantify the performance of the trained model by the accuracy based on data from the validation partition. Preliminary experiments with SlimCNN \cite{sharma_slim-cnn_2020} didn't produce satisfying accuracies. EfficientNet is an alternative convolutional network \cite{tan2019efficientnet} that can scale the depth, width, and resolution of all filters with a single parameter (we use EfficientNet-B1). Since UTKFace does not have any partitioning information, we split the data set randomly into train, validation, and test partitions which contain $70\%$, $20\%$, and $10\%$ of the samples. 

\begin{table*}[!t]
	\centering
	\caption{Quantitative outcomes with manually selected weighting coefficients $\lambda$ for predicting the multi-class target attribute $y_t = Age\,group$ on the UTKFace facial image data set. Experiments \#1 to \#6 concern the sensitive attribute $y_s = Gender$ and \#7 to \#12 cover the sensitive attribute $y_s = Ethnicity$.}
	\vspace{5pt}
	\begin{tabular}{|*{8}{c|}}
		\hline
		\scriptsize\# & Loss           & $Acc$   &   $L_{iou}$ \eqref{eq:iou_losssqrt}  & $L^{l_2}_{eo}$ \eqref{eq:eo_losssqrt} & $L^{mi}_{eo}$ \eqref{eq:eo_lossmi} & $L^{l_2}_{dp}$ \eqref{eq:dp_losssqrt} & $L^{mi}_{dp}$ \eqref{eq:dp_lossmi}  \\ \hline
		\scriptsize1  & $L_{ce}$       & 0.847   & \num{1.45E-03} & \num{1.66E-01} & \num{1.23E-01} & \num{6.55E-02} & \num{4.21E-02} \\
		\scriptsize2  & $L_{iou}$      & 0.852   & $\mathbf{3.04 \times 10^{-4}}$ & \num{3.00E-02} & \num{2.89E-02} & \num{3.77E-02} & \num{2.18E-02} \\
		\scriptsize3  & $L^{l_2}_{eo}$ & 0.856   & \num{3.08E-03} & \num{8.11E-02} & \num{7.21E-02} & \num{4.85E-02} & \num{3.10E-02} \\
		\scriptsize4  & $L^{mi}_{eo}$  & $\mathbf{0.857}$ & \num{9.82E-04} & $\mathbf{2.88 \times 10^{-2}}$ & $\mathbf{2.25 \times 10^{-2}}$ & \num{2.16E-02} & \num{1.36E-02} \\
		\scriptsize5  & $L^{l_2}_{dp}$ & 0.852   & \num{5.61E-04} & \num{3.59E-02} & \num{3.59E-02} & \num{9.39E-03} & \num{7.00E-03} \\
		\scriptsize6  & $L^{mi}_{dp}$  & 0.848   & \num{7.78E-04} & \num{9.27E-02} & \num{6.40E-02} & $\mathbf{5.33 \times 10^{-3}}$ & $\mathbf{4.51 \times 10^{-3}}$ \\ \hline
		\scriptsize7  & $L_{ce}$       & 0.846   & \num{2.06E-02} & \num{3.73E-01} & \num{1.81E-01} & \num{1.96E-01} & \num{5.48E-02} \\
		\scriptsize8  & $L_{iou}$      & 0.847   & \num{6.62E-03} & $\mathbf{9.75 \times 10^{-2}}$ & \num{8.60E-02} & \num{1.98E-01} & \num{4.53E-02} \\
		\scriptsize9  & $L^{l_2}_{eo}$ & 0.844   & \num{1.85E-02} & \num{2.50E-01} & \num{1.53E-01} & \num{1.91E-01} & \num{4.70E-02} \\
		\scriptsize10 & $L^{mi}_{eo}$  & $\mathbf{0.857}$ & \num{1.53E-02} & \num{1.12E-01} & \num{1.08E-01} & \num{1.52E-01} & \num{4.21E-02} \\
		\scriptsize11 & $L^{l_2}_{dp}$ & $\mathbf{0.857}$ & \num{1.77E-02} & \num{1.48E-01} & \num{1.62E-01} & \num{8.85E-02} & \num{3.03E-02} \\
		\scriptsize12 & $L^{mi}_{dp}$  & 0.854   & $\mathbf{6.61 \times 10^{-3}}$ & \num{1.03E-01} & $\mathbf{5.19 \times 10^{-2}}$ & $\mathbf{3.26 \times 10^{-2}}$ & $\mathbf{6.24 \times 10^{-3}}$ \\ \hline
	\end{tabular}
	
	\label{tab:utkface}
\end{table*}

Table \ref{tab:utkface} shows the quantitative results from the experiments with UTKFace. The interpretation of the rows and columns is the same as in Table \ref{tab:celeba}. Again, the model fairness improved w.r.t.~to almost all proposed fairness metrics for both sensitive attributes. In addition, applying any fairness loss led to an improvement in the prediction accuracy with the sensitive attribute $y_s = Gender$. Experiments with the sensitive attribute $y_s = Ethnicity$ also improved the prediction accuracy except when we applied the $L^{l_2}_{eo}$ loss.

%\paragraph{\normalfont\textbf{SIIM-ISIC Melanoma Classification}}
\subsection{SIIM-ISIC Melanoma Classification}
\label{sec:mela}
The prediction target attribute in our experiments is a diagnosis, whether a lesion is malignant or benign. Each image has annotations of a binary gender variable (Male, Female) which we use as the sensitive attribute. The performance of the trained model is quantified with the area under the receiver operating curve ($AUC$) on the validation partition, which was the standard evaluation metric in this challenge. We use EfficientNet-B1 as the classification model in all experiments as with UTKFace. As the data set only contains annotations in the original train partition, we used these annotated images and randomly assigned them to train, validation, and test partitions consisting of $70\%$, $20\%$, and $10\%$ of the original train samples. We used different transformations to augment the training data, which improved the baseline $AUC$ score (details in the supplementary material). Since the data set only contains a small fraction of malignant samples, we used the effective number of samples \cite{cui2019class} as a class weighting method to deal with this label imbalance. Each sample is assigned a normalized weight $\alpha_i = \frac{1 - \beta}{1 - \beta^{n_i}}$ according to the frequency $n_i$ of the $i$-th class in the train partition. The hyper-parameter $\beta$ adjusts these weights according to the label distribution in a particular data set, which we set to $\beta = 0.9998$. Table \ref{tab:siim_isic_melanoma} shows the results of the experiments with the fine-tuned class-weighting. The application of the EO-based fairness losses $L^{l_2}_{eo}$ and $L^{mi}_{eo}$ improved the $AUC$ score considerably. 
%%%%%%% add rebuttal 5 %%%%%  very weak statement not highlighting bad results and/or mentioning good results i.e. IOU
Additionally, incorporation of our novel $L_{iou}$ based fairness loss helped to improve the fairness of the model w.r.t the baseline for all of the proposed fairness-based evaluation metrics. 

\begin{table*}[!t]
	\centering
	\caption{Experimental results with manually selected weighting coefficients $\lambda$ for predicting the binary target attribute $y_t = Diagnosis$ from skin lesion images with the SIIM-ISIC melanoma classification data set. The experiments use the binary variable $Gender$ as the sensitive attribute $y_t$.}
	\vspace{5pt}
	\begin{tabular}{|*{8}{c|}}
		\hline
		\scriptsize\# & Loss           & $AUC$ &   $L_{iou}$ \eqref{eq:iou_losssqrt}  & $L^{l_2}_{eo}$ \eqref{eq:eo_losssqrt} & $L^{mi}_{eo}$ \eqref{eq:eo_lossmi} & $L^{l_2}_{dp}$ \eqref{eq:dp_losssqrt} & $L^{mi}_{dp}$\eqref{eq:dp_lossmi}  \\ \hline
		\scriptsize1  & $L_{ce}$       & 0.829 & \num{1.26E-03} & \num{6.22E-02} & \num{3.51E-02} & \num{3.54E-04} & \num{9.27E-04} \\
		\scriptsize2  & $L_{iou}$      & 0.801 & $\mathbf{5.71 \times 10^{-5}}$ & $\mathbf{3.17 \times 10^{-2}}$ & $\mathbf{1.67 \times 10^{-4}}$ & $\mathbf{1.67 \times 10^{-4}}$ & $\mathbf{1.32 \times 10^{-4}}$ \\
		\scriptsize3  & $L^{l_2}_{eo}$ & $\mathbf{0.869}$ &\num{4.10E-04} & \num{8.52E-02} & \num{3.95E-02} & \num{7.60E-04} & \num{1.27E-03} \\
		\scriptsize4  & $L^{mi}_{eo}$  & 0.854 & \num{7.54E-04} & \num{7.71E-02} & \num{4.38E-02} & \num{7.41E-04} & \num{1.43E-03} \\
		\scriptsize5  & $L^{l_2}_{dp}$ & 0.804 & \num{6.62E-04} & \num{8.44E-02} & \num{4.86E-02} & \num{3.53E-04} & \num{8.89E-04} \\
		\scriptsize6  & $L^{mi}_{dp}$  & 0.835 & \num{3.64E-04} & \num{4.71E-02} & \num{2.39E-02} & \num{4.52E-04} & \num{9.43E-04} \\ \hline
	\end{tabular}
	\label{tab:siim_isic_melanoma}
\end{table*}

%%%%%%%%%%%% add rebuttal 7 %%%%%%%%%%%%%%
It is to be noted that our work is not comparable to  closely related approaches. Some works \cite{ref98,ref99} propose to remove biases w.r.t. sensitive attributes from the feature representation of the model. Our approach instead focuses on enabling the prediction accuracy of the target attribute to not depend on sensitive attributes. %Please note that \cite{ref98, ref99} with their unique tasks have no comparison to previous work but comprehensive parametric studies instead.
Recent works with similar tasks as ours propose loss functions based on distance \cite{ref106} or cosine similarity \cite{ref107} measures while we explore the inherent fairness-driven probabilistic measures as the loss functions in our experimental setup. Such loss functions were not studied before for visual data sets, so a reasonable comparison with methods such as \cite{ref106,ref107} is also not possible.

%\paragraph{\normalfont\textbf{Accuracy-Fairness Trade-Off}}
%\subsection{Accuracy-Fairness Trade-Off}
\subsection{Effect of $\lambda$ on Fairness vs Accuracy}
\label{sec:lambda}
\begin{figure}[!t]
    \centering      
    \includegraphics[width=\textwidth]{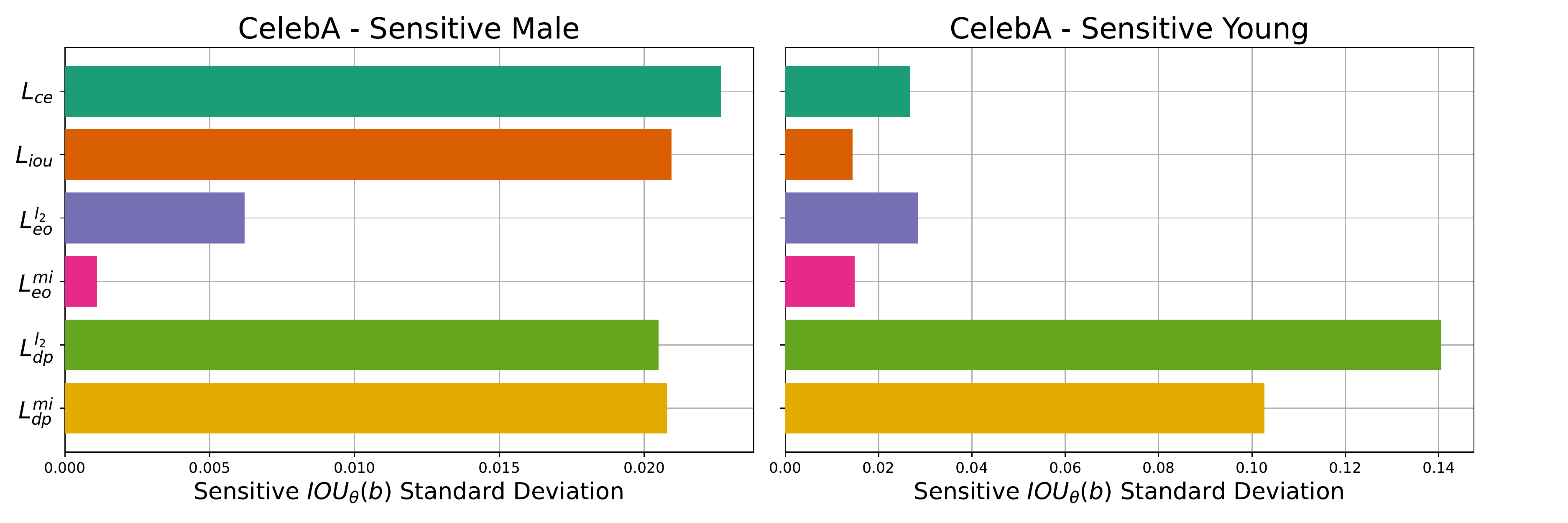}
    \caption{
        Experimental results with the fairness loss weighting coefficient $\lambda$. The standard deviation of the $IoU_{\theta}(b)$ for different sensitive class labels $b$ is used to quantify the model fairness. The prediction accuracy quantifies the classification performance.
    }
    \label{fig:sensitive_iou_chart}
\end{figure}
%%%%%%%%%%%% add rebuttal 1 & 2 %%%%%%%%%%%%%%
We studied the effect of the coefficient $\lambda$ on both model fairness and classification accuracy. Its value depends on the dataset, optimized loss, and whether achieving high accuracy is more relevant for a specific application or fairness. 
%We believe that by fixation of $\lambda$ for a particular use case, our approach would apply to a wide range of problems.
Note that setting $\lambda=0$ reverts to the original image classification problem.
%Hence, a suitable choice of $\lambda$ can not worsen the results in principle.  
We obtained the results for the initial experiments by heuristically searching for good weighting coefficients. We observed that starting with small $\lambda$-values is beneficial and studied whether the obtained model is fair. One could increase $\lambda$ until the classification performance significantly decreases. Later, we performed hyper-parameter optimization (HPO) to find $\lambda$ that leads to the best improvement in the model fairness with maintaining high classification accuracy.  

\paragraph{\normalfont\textbf{Heuristic Search}}  We use the Bessel corrected standard deviation $\sigma_{IoU}(\lambda) = \sqrt{\frac {1}{K_s - 1}\sum _{i=1}^{K_s}(IoU_{\theta}(b_i) - \overline{IoU_{\theta}(b)})^2}$
of the $IoU_{\theta}(b)$ \eqref{eq:iou2} as a fairness evaluation measure and the prediction accuracy as a performance measure. The results for the manual selection of $\lambda$ on CelebA can be seen in Fig.~\ref{fig:sensitive_iou_chart}. The $L_{iou}$ and $L_{eo}^{mi}$ losses improved the fairness w.r.t.~the baseline models (trained with the standard cross-entropy loss $L_{ce}$) for both sensitive attributes. In contrast, the $L_{eo}^{l_2}$ loss and the DP-based losses $L_{dp}^{mi}$ and $L_{dp}^{mi}$ improved the fairness only for the sensitive attribute $y_s = Male$. 

\paragraph{\normalfont\textbf{Hyperparameter Optimization (HPO)}}

\begin{figure}[!t]
    \centering      
    \includegraphics[width=.32\textwidth]{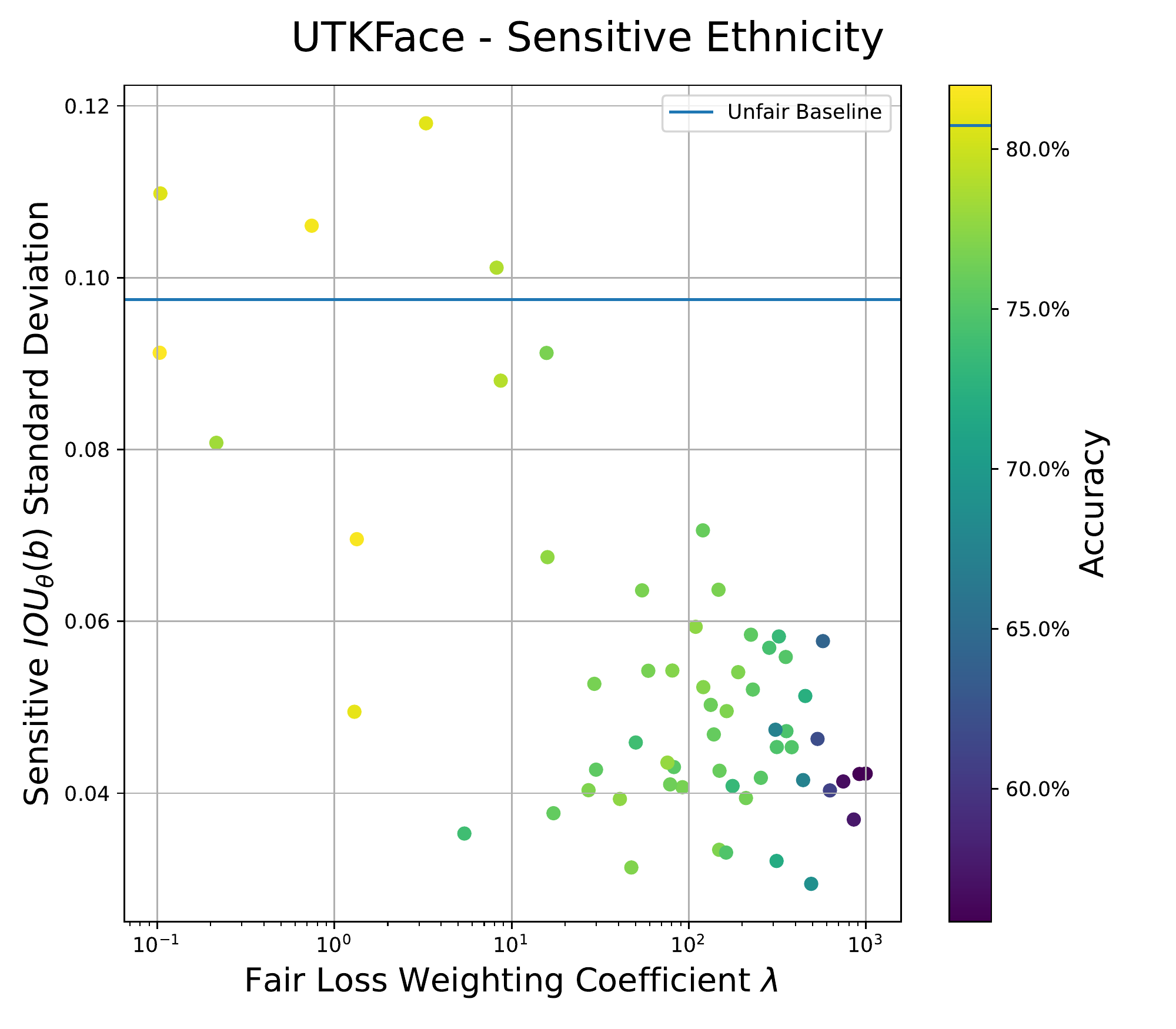}
    \includegraphics[width=.32\textwidth]{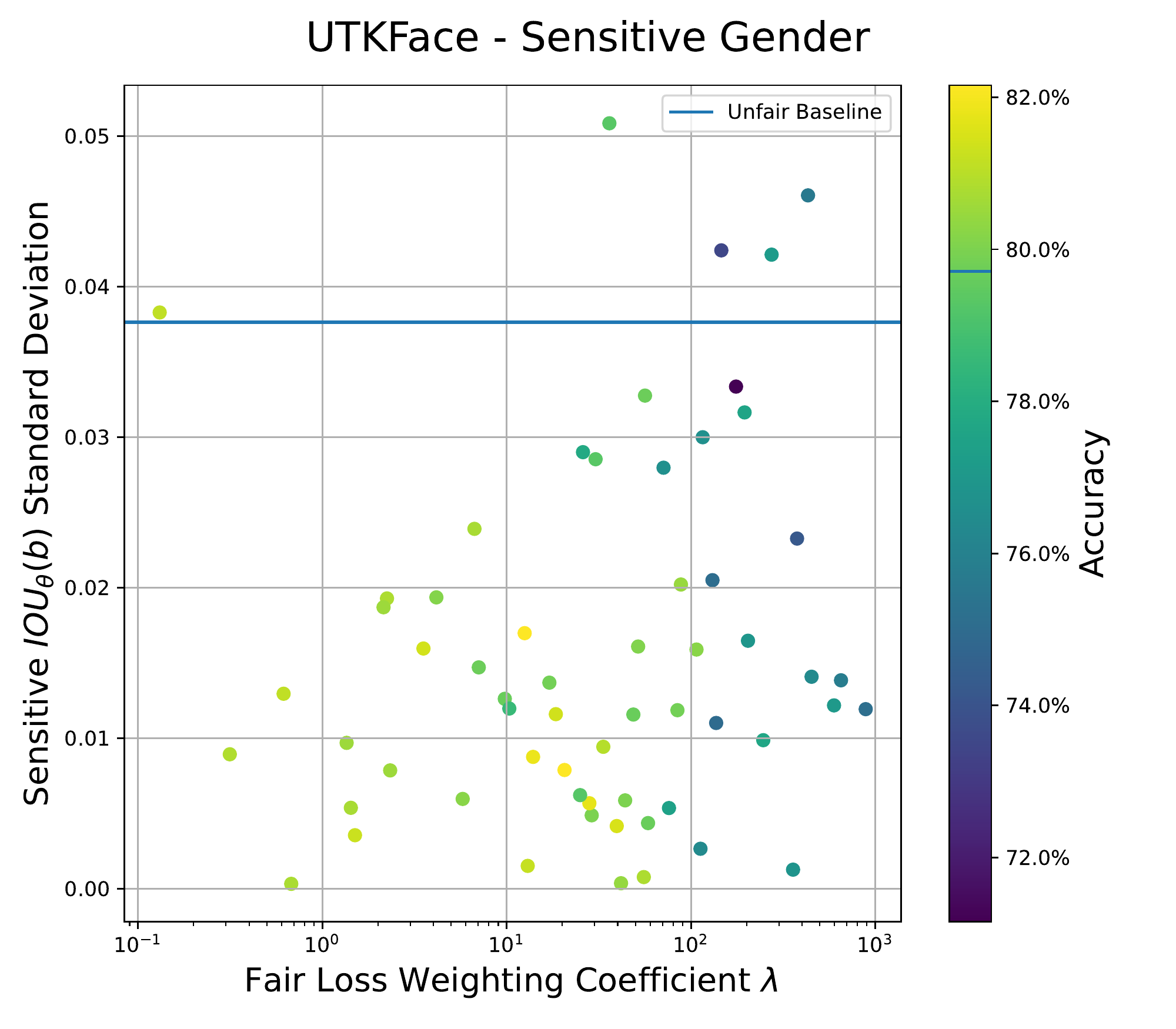}
    \includegraphics[width=.32\textwidth]{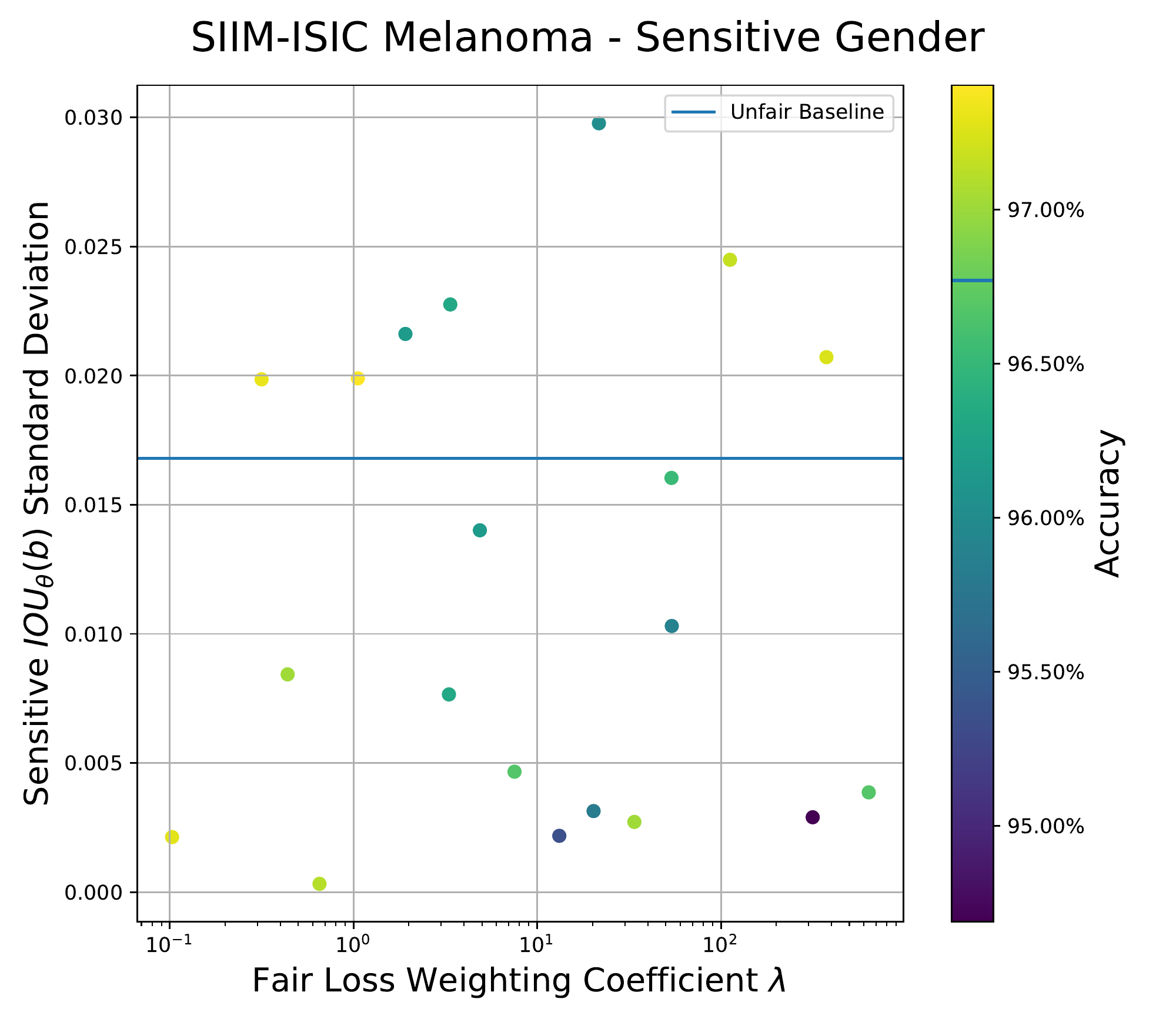}
    \caption{
        Experimental results with the fairness loss weighting coefficient $\lambda$ on UTKFace and SIIM-ISIC Melanoma. The standard deviation of the $IoU_{\theta}(b)$ for different sensitive class labels $b$ is used to quantify the model fairness (less is fairer). The prediction accuracy quantifies the model performance. The blue lines show the baseline performance and fairness values.
    }
    \label{fig:hpo_results}
\end{figure}

We performed HPO of the weighting coefficient $\lambda$ for the UTKFace and SIIM-ISIC Melanoma data sets with $\sigma_{IoU}(\lambda)$ as the minimization objective and searching $\lambda$ within $[\num{1e-01}, \num{1e+03})$. We use the validation partition for the HPO trials and evaluate the resulting models on the test partition. Due to time and resource constraints, we restrict our experiments to the novel IoU loss $L_{iou}(\theta)$ \eqref{eq:iou_losssqrt} with 60 HPO trials on the UTKFace data set and 20 trials on the SIIM-ISIC Melanoma classification data set. The results of these HPO experiments are shown in Fig.~\ref{fig:hpo_results}. We observe a clear trend for experiments on UTKFace with the sensitive attribute $y_s = Ethnicity$ (left figure), where the fairness of the baseline model is relatively low. With an exponentially increasing $\lambda$, there is a linear improvement in the model fairness and a linear decrease in the prediction accuracy.
%That represents a classical trade-off relationship between these two evaluation metrics. 
However, such a trend does not exist for experiments with UTKFace for the sensitive attribute $y_s = Gender$ (figure in the middle), where the baseline fairness is already good. Here, the model fairness and the prediction accuracy decrease linearly with exponentially increasing $\lambda$ values. Additionally, there is more variation in the fairness improvements when $\lambda$ is larger. In these experiments with UTKFace, there is a certain range of $\lambda$ values (the region with yellow dots) that simultaneously improves the model fairness and prediction accuracy. The baseline fairness is already high for the SIIM-ISIC Melanoma data set with the sensitive attribute $y_s = Gender$ (right figure). Hence, the prediction accuracy and fairness improvement are independent of the $\lambda$ value, as its effect on these metrics seems random.

\section{Conclusion}

In this work, we presented the theoretical intuition toward obtaining fair image classification models. We implemented various fairness metrics as standardized differentiable loss functions for categorical variables and compared their effectiveness in bias mitigation when compared to our novel IoU loss. Our experiments on publicly available facial and medical image data sets show that the proposed fairness losses do not degrade the classification performance on target attributes and reduce the classification bias at the same time. With this work and the publication of our source code, we provide a tool that encourages further work in this research direction. An interesting future work would be the visualization of relevant regions in the input image space that make the fair model less biased compared to the baseline model, trained with a standalone cross-entropy loss.

\paragraph{\bf Acknowledgement.} 
This work primarily received funding from the German Federal Ministry of Education and Research (BMBF) under \textit{Software Campus} (grant 01IS17044) and the Competence Center for Big Data and AI \textit{ScaDS.AI Dresden/Leipzig} (grant 01/S18026A-F).
 The work also received funding from Deutsche Forschungsgemeinschaft (DFG) (grant 389792660) as part of TRR 248 and the Cluster of Excellence \textit{CeTI} (EXC2050/1, grant 390696704). The authors gratefully acknowledge the Center for Information Services and HPC (ZIH) at TU Dresden for providing computing resources. 
 %Finally, the authors would like to thank all the anonymous reviewers for providing helpful suggestions. 

\bibliographystyle{unsrtnat}
\bibliography{ms}

\end{document}